\documentclass[sigconf]{acmart}
\usepackage{graphicx}
\usepackage{threeparttable}
\usepackage{balance}

\AtBeginDocument{%
  }

\copyrightyear{2024}
\acmYear{2024}
\setcopyright{acmlicensed}
\acmConference[UbiComp Companion '24] {Companion of the 2024 ACM International Joint Conference on Pervasive and Ubiquitous Computing}{October 5--9, 2024}{Melbourne, VIC, Australia.}
\acmBooktitle{Companion of the 2024 ACM International Joint Conference on Pervasive and Ubiquitous Computing (UbiComp Companion '24), October 5--9, 2024, Melbourne, VIC, Australia}
\acmISBN{979-8-4007-1058-2/24/10}
\acmDOI{10.1145/3675094.3680520}

\acmSubmissionID{1007}

\begin{document}

\title{GCCRR: A Short Sequence Gait Cycle Segmentation Method Based on Ear-Worn IMU}

  

\author{Zhenye Xu}
\email{lingyy@sjtu.edu.cn}
\orcid{0000-0001-9449-4846}
\affiliation{%
    \institution{Shanghai Jiao Tong University}
    \department{Institute of Medical Robotics, School of Biomedical Engineering}
    \city{Shanghai}
    \country{China}
}

\author{Yao Guo}
\authornote{Corresponding author}
\email{yao.guo@sjtu.edu.cn}
\orcid{0000-0001-8041-1245}
\affiliation{%
    \institution{Shanghai Jiao Tong University}
    \department{Institute of Medical Robotics, School of Biomedical Engineering}
    \city{Shanghai}
    \country{China}
}


\begin{abstract}
This paper addresses the critical task of gait cycle segmentation using short sequences from ear-worn IMUs, a practical and non-invasive approach for home-based monitoring and rehabilitation of patients with impaired motor function. While previous studies have focused on IMUs positioned on the lower limbs, ear-worn IMUs offer a unique advantage in capturing gait dynamics with minimal intrusion. To address the challenges of gait cycle segmentation using short sequences, we introduce the Gait Characteristic Curve Regression and Restoration (GCCRR) method, a novel two-stage approach designed for fine-grained gait phase segmentation. The first stage transforms the segmentation task into a regression task on the Gait Characteristic Curve (GCC), which is a one-dimensional feature sequence incorporating periodic information. The second stage restores the gait cycle using peak detection techniques. Our method employs Bi-LSTM-based deep learning algorithms for regression to ensure reliable segmentation for short gait sequences. Evaluation on the HamlynGait dataset demonstrates that GCCRR achieves over 80\% Accuracy, with a Timestamp Error below one sampling interval. Despite its promising results, the performance lags behind methods using more extensive sensor systems, highlighting the need for larger, more diverse datasets. Future work will focus on data augmentation using motion capture systems and improving algorithmic generalizability. 
\end{abstract}



\begin{teaserfigure}
\centering
  \includegraphics[width=1\textwidth]{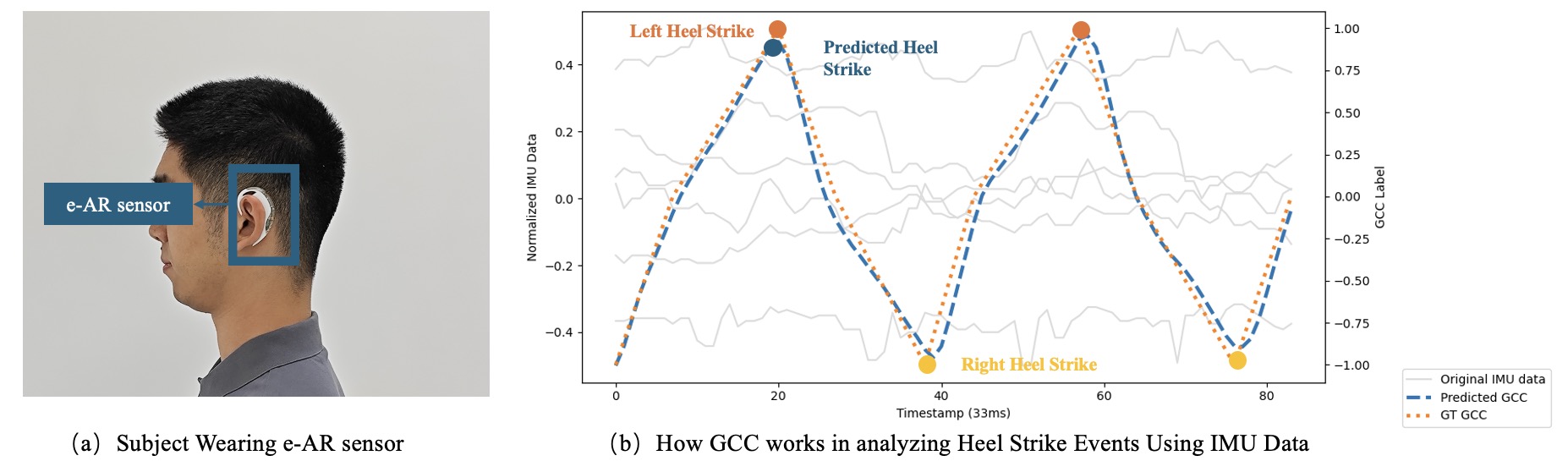}
  \caption{Overview of Our Work on Gait Cycle Segmentation Using Ear-Worn IMU}
  \label{fig:teaser}
\end{teaserfigure}


\maketitle

\section{Introduction}
Quantitative analysis of human gait can help assess the motor abilities of patients with impaired motor function and assist doctors in providing rehabilitation guidance and suggestions. This analysis can be conducted using various sensors, with IMU-based methods offering unique advantages. Compared to vision-based methods, IMU-based methods are easier to deploy, offer more consistent wearability, and pose a lower risk of privacy leakage\cite{guo2022detection, aziz2023earables}. Thus, IMU-based methods make it possible to monitor and analyze abnormal gait at home-based environments.

Many studies have explored IMU-based gait analysis methods, with IMUs worn on the waist, knee joint, or shoes\cite{zhou2020validation, mazilu2013engineers, cola2015node}. However, research by Atallah et al.\cite{atallah2011sensor} indicates that ear-worn IMUs can sensitively capture changes in body posture during walking, and Min et al.'s research\cite{min2018exploring} has also validated the potential of ear-worn IMU in kinetic sensing.

Nowadays, quantitative analysis is primarily achieved through gait parameter extraction. Key gait parameters include gait speed, step length, cadence, and more\cite{guo2022detection}. Precise gait cycle segmentation from the original sensor signals is essential to obtain these parameters. Many studies focus on gait cycle segmentation using lower limb-positioned IMUs. However, few studies have investigated gait segmentation methods based on ear-worn IMUs. 
Jarchi et al. proposed a gait segmentation method that relies on the singular value decomposition (SVD) of input signals and validated its performance on Parkinson's Disease (PD) patients\cite{jarchi2014validation}. Diao et al. proposed improvements to this method\cite{diao2020novel}. However, both methods rely on sufficiently long periods of straight-line walking, which can present spatial challenges in a home environment. 
Prakash et al. presented another method to count users' steps that is robust across different walking patterns and users based on ear-worn IMU\cite{prakash2019stear}. However, this method does not focus on patients with abnormal gait and only performs gait counting without providing temporal markers for gait cycle segmentation, thus not supporting fine-grained gait cycle segmentation.

In home environments, conducting gait data collection and segmenting gait cycles faces a significant challenge, notably due to limited space. 
While we can categorize various motion patterns using ear-worn devices, this approach may introduce bias in subsequent abnormal gait analysis procedures.
This is especially difficult for urban residents, where cramped spaces make it hard to independently complete more than five gait cycles in a straight-line walk. 
For long-term rehabilitation patients with impaired mobility, completing several gait cycles in one go is also challenging. 
Therefore, segmenting gait cycles using short sequences is a critical technological bottleneck. 
As mentioned above, current methods using ear-worn sensors for gait cycle segmentation \cite{jarchi2014validation, prakash2019stear} are inadequate for short sequences with fewer cycles or less evident periodicity. Therefore, there is a need for a method to segment gait cycles based on short sequences.

To address these limitations, we propose a novel approach called Gait Characteristic Curve Regression and Restoration (GCCRR). This method involves two stages: the first stage transforms the fine-grained gait phase segmentation task into a regression task on the Gait Characteristic Curve (GCC). By doing so, we can explicitly incorporate periodic information into the segmentation process. The second stage restores the gait cycle based on the predicted GCC using peak detection techniques. By incorporating LSTM-based deep learning algorithms, this method ensures reliable segmentation results while maintaining compatibility with short sequences. With the short sequence gait cycle segmentation method, we could perform real-time collection and analysis of patients' gait sequences based on their recently produced gait cycles.

The remainder of this paper is structured as follows: Section 2 introduces the ear-worn IMU-based gait segmentation, providing an overview of the approach, describing the devices used, detailing the Gait Characteristic Curve (GCC), and explaining the GCC Regression and Restoration process. Section 3 focuses on the evaluation of the proposed method, covering the dataset, model training, evaluation metrics, and result analysis. Section 4 presents a discussion of the findings. Section 5 concludes the paper and outlines directions for future research.
 
\section{Ear-worn IMU based Gait Segmentation
}

\subsection{Overview}
The key to segmenting gait cycles based on IMU sequences is to accurately identify the heel-strike event while walking. This is usually defined as a classification task for each time point across the whole input sequence. In this work, we take a straight walking IMU sequence with limited length as input and give out a classification label for each time point. 

\subsection{Ear-Worn IMU Devices}

The acquisition of data is facilitated by the e-AR sensor developed by the ICL group. As described in \cite{atallah2013ear,jarchi2014validation}, the sensor is mainly composed of an 8051 processor, equipped with a 2.4GHz transceiver (Nordic nRF24E1), a 3D accelerometer (Analog Devices ADXL330), a 2MB EEPROM (Atmel AT45DB161), and a 55mAhr Li-Polymer battery. This sensor is connected to a host computer via Bluetooth Low Energy (BLE) technology when collecting data and the sampling rate is 30Hz. 

Considering the electromagnetic interference of most ear-worn devices\cite{kawsar2018earables}, we use only the six-axis IMU data, which includes accelerometer and gyroscope data, in our subsequent analysis. Previous studies have validated the reliability of this sensor in gait analysis, and its consistency across repeated wearings ensures effective continuous tracking and data collection.

\subsection{Gait Characteristic Curve (GCC)}
As we all know, while people walk along a straight line, their gait data exhibit a certain periodicity due to the alternating activity of both feet. The periodicity can be easily obtained from long walking sequences through either time-domain or frequency-domain characteristics. However, it may fail in short walking sequences with less than three gait cycles, which are easy for people to collect by themselves at home but contain less obvious periodic information. Once segmentation is performed without periodicity, the resulting classification may exhibit irregular segmentation labels, which are deemed unacceptable.

Our key motivation is to explicitly provide the algorithm with the current gait phase information and then use specific methods to reconstruct the gait cycle. Thus, we can turn the classification task into a two-step task consisting of GCC regression and gait phase restoration.

GCC is the signal we construct containing periodic information of gait data. As shown in Figure \ref{fig:teaser}, we set the left-heel-strike events of the GCC to 1 and the right-heel-strike events to -1, while toe-off events are all set to 0. Linear interpolation is then applied to fill in all the missing points. Thus we can draw a curve that can provide relative value in the gait cycle for each timestamp.

\begin{figure}
    \includegraphics[scale=0.55]{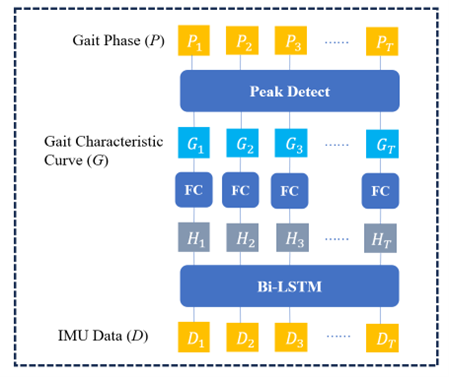}
    
    \caption{Overview of the GCCRR method}
    \label{fig:GCCRR-framework}
\end{figure}

\subsection{GCC Regression and Gait Cycle Restoration}
As shown in Figure \ref{fig:GCCRR-framework}, by performing GCC regression based on Bi-LSTM and gait cycle restoration based on peak detection, we can provide a method for gait cycle segmentation based on short-time sequences.

The GCC Regression is primarily based on the Bi-LSTM framework\cite{schuster1997bidirectional} and following fully connected layers. The original IMU data is fed into a 5-layer Bi-LSTM network, and the output hidden state \(H_i\) for each timestamp is passed through a 2-layer fully connected neural network to get the predicted GCC \(G_i\).

To restore the gait cycle from the predicted GCC sequence, a peak detection method provided by \cite{jiang2017robust} is applied to the predicted GCC sequence. The local highest peak of the predicted GCC sequence \(G\) is treated as the left-heel-strike event, while the local lowest peak is treated as the right-heel-strike event. This approach enables gait cycle segmentation based on the heel-strike events. 

It is worth mentioning that as we treat the task as a classification task for each timestamp if we treat the heel strike event as one class while others as another class, simply using the classification accuracy to measure the performance of the algorithm for all time points will not work well since the classification task is highly imbalanced across classes. To solve this, we try to convert the output sequence to more class-balanced ones. So here we introduce the Gait Phase \(P\) as the segmentation output, where we set \(P_i=0\) for timestamps from left heel strike to right and \(P_i=1\) for timestamps from right heel strike to left. Each gait cycle contains a \(P_i=0\) part and a \(P_i=1\) part, and the class label is more balanced to apply evaluation metrics.

\section{Experiments and Results}

\begin{table}[]
    \setlength{\abovecaptionskip}{0.235cm}
    \setlength{\belowcaptionskip}{-0.1cm}
    \caption{The total amount of data in various modalities in the HamlynGait dataset}
    \centering
    \scalebox{0.8}{
    \begin{tabular}{ccccccc}
        \toprule
        Subject & e-AR Wear Side & Normal & Toe-in & Toe-out & Supination & Pronation \\
        \midrule
        \#1     & L              & 58     & 48     & 51      & 22         & 27        \\
        \#2     & L              & 24     & 26     & 29      & 38         & 31        \\
        \#3     & L              & 24     & 36     & 40      & 23         & 38        \\
        \#4     & L              & 12     & 20     & 23      & 21         & 24        \\
        \#5     & R              & 25     & 33     & 31      & 34         & 34        \\
        \#6     & R              & 25     & 24     & 23      & 26         & 44        \\
        \#7     & R              & 24     & 30     & 27      & 47         & 56        \\
        \#8     & R              & 24     & 0      & 0       & 32         & 0         \\
        \#9     & R              & 26     & 20     & 39      & 25         & 29        \\
        \#10    & R              & 25     & 47     & 43      & 46         & 48        \\
        \#11    & R              & 16     & 11     & 13      & 11         & 15        \\
        \#12    & R              & 5      & 0      & 0       & 17         & 12        \\
        \#13    & R              & 7      & 6      & 12      & 7          & 13         \\
    \bottomrule
    \end{tabular}
    }
    \label{tab:HamlynGait}
    \vspace{-5pt}
\end{table}

\subsection{Dataset}
To facilitate the training and evaluation of our LSTM-based model, we utilized the Hamlyn Gait dataset\cite{guo2021mcdcd}, which comprises ear-worn IMU data and motion capture data from both typical individuals and individuals mimicking abnormal gait patterns, including toe-in, toe-out, supination and pronation. This dataset can provide the gold standard timestamps for heel strike and toe-off events based on the data obtained through a motion capture system. During data collection, participants were instructed to walk back and forth within a 3×3-meter space. The dataset encompasses data from 13 subjects, encompassing 5 distinct gait modalities. Table \ref{tab:HamlynGait} presents the number of gait cycles observed for each subject and gait mode. It is worth noting that each straight walking period usually contains 1 to 3 gait cycles.

\subsection{Implementation Details}
We quantitatively assessed our model's performance in gait cycle segmentation using the HamlynGait dataset. These experiments were carried out on a workstation running Ubuntu 16.0 and equipped with an NVIDIA RTX 2080 GPU. Our Bi-LSTM-based model was implemented using Python 3.8 and PyTorch 2.0.0. We set the learning rate for the Bi-LSTM model at 0.0008, and the size of the hidden state is set to 500. 

The model's training utilized acceleration and angular velocity data from all walking gait patterns, encompassing various individuals and gait modes, to establish a comprehensive and robust prediction system. Recognizing that each subject exhibits distinct gait characteristics, we employed the Leave One Subject Out (LOSO) methodology to validate the model's performance. Specifically, we isolated all data from a single subject as the test set, partitioned the remaining data into training and validation sets at a 9:1 ratio, and subsequently trained the Bi-LSTM model by minimizing the Mean Square Error (MSE) on the training set. It is worth mentioning that we trained two separate models based on the wearing side of the e-AR sensor, one for the left side and one for the right side.

To adapt to various step lengths of different individuals, we introduced a random time-stretching technique for the gait data used during training, with stretching factors ranging from 1 to 4 times. This augmentation process was performed four times for each dataset. As a result, the entire training dataset was augmented five times, ensuring the model's adaptability to cadence variations.

\subsection{Evaluation Metrics}
When it comes to evaluating the cycle segmentation algorithms, previous works focus on extracting gait parameters such as cadence, step length, and so on. However, due to the limited cycles contained in the short sequences, new evaluation parameters should be provided in this task. Here we come up with three quantitative parameters to assess the model's proficiency and resilience in extracting gait cycles, which are the Accuracy, the False Peak Rate, and the Timestamp Error. The details of the three parameters will be shown below.

The Accuracy is defined as the ratio of the number of correctly predicted \(P_i\)  to the total number of \(P_i\) across the dataset. It can reflect the overall accuracy of the gait cycle segmentation algorithm.

Another metric that we are concerned about is the False Peak Rate. The False Peak Rate is the ratio of the number of sequences in which heel strike events are incorrectly detected (less or more in number) to all the number of sequences. It can reflect the failure caused by the incorrectly identified heel strike events.

In contrast, the Timestamp Error can reflect the failure caused by the timestamp shift of the successfully detected heel strike events. It is calculated by averaging the timestamp difference across all the successfully detected heel strike events.

\begin{table}[]
    \setlength{\abovecaptionskip}{0.235cm}
    \setlength{\belowcaptionskip}{-0.1cm}
    \caption{Classification results of GCCRR on HamlynGait dataset}
    \begin{tabular}{cccc}
        \toprule
        Subject & Accuracy (\%) & False Peak Rate (\%) & Timestamp Error (s)\\
        \midrule
        Left*    & 88.4  &   6.78 &   0.0065   \\
        Right   & 80.7  &   7.85 &   0.0199   \\
    \bottomrule
    \end{tabular}

    \label{tab:results}

\end{table}

\begin{table}[]
    \setlength{\abovecaptionskip}{0.235cm}
    \setlength{\belowcaptionskip}{-0.1cm}
    \caption{Ablation study of GCCRR}
    \begin{tabular}{cccc}
        \toprule
         & Left Accuracy(\%) &  Right Accuracy(\%) \\
        \midrule
        Bi-LSTM+FC(w/o GCC)   & 81.4  &   65.6  \\
        GCCRR    & 88.4  &   80.7    \\

    \bottomrule
    \end{tabular}

    \label{tab:ablation}

\end{table}

\subsection{Result Analysis}

The performance of our GCCRR method is presented in Table \ref{tab:results}. All the three metrics are averaging across the LOSO experiments. As described above, our model is training on two subject sets, divided by the wearing side of the ear-worn sensor, so the measurement is done respectively, too. The Left subject set includes all individuals who wore a sensor on their left ear during data collection, while the Right subject set comprises those who wore a sensor on their right ear. We found that the accuracy of GCCRR on both subject sets is above 80\% overall. Furthermore, in both subject sets, the average Timestamp Error falls below one sampling interval, approximately 0.033 seconds (30Hz). This indicates that the error is primarily due to over-detected or missing peaks rather than discrepancies in timestamps between the predicted and actual peaks.

To demonstrate the necessity and effectiveness of introducing GCC, we used a Bi-LSTM + FC structure to directly predict \( P_i \) to compare with our proposed GCCRR. The results are shown in Table \ref{tab:ablation}, highlighting that the extraction of GCC can significantly improve the performance of the task, with the improvement by 7.0\% and 15.1\% of both subject sets, respectively.

When comparing the Left and Right subject sets, GCCRR performs better on the Left subject set, which contains fewer subjects, and the training data of all the subjects are sufficient. This result may indicate a lack of generalization in the model, which is one of our key areas for future improvement. However, compared with the model without GCC, the Accuracy on the Right subject set gains more than on the Left, which means that the GCC may help more while the data is insufficient.

\section{Discussion}
With GCC providing gait cycle information, the performance of the gait cycle segmentation algorithm has significantly improved. However, the accuracy is still not as satisfying compared to gait cycle segmentation methods based on other sensor systems, such as motion capture systems or various IMU systems worn on the lower limb. It is worth noting that the gait cycle segmentation algorithm based on ear-worn IMUs is still under development, especially for short sequences with limited cycles. In our view, the bottleneck of the problem mainly lies in the lack of ear-worn IMU datasets that contain gold standard heel strike events. Ear-worn IMUs inherently have a higher level of noise, necessitating more extensive data to support deep learning-based gait cycle extraction algorithms. The generation of IMU signals based on motion capture systems or videos is drawing considerable attention. This approach may greatly assist with data augmentation in future work.

\section{Conclusion}
This paper emphasizes the importance of gait segmentation based on short ear-worn IMU sequences. It introduces a gait cycle segmentation method, GCCRR, which is well-suited for short sequence gait cycle segmentation and proposes three evaluation metrics for this task. To the best of our knowledge, GCCRR is the first algorithm designed to address this specific task, and it achieves relatively good performance. To further this work, we will focus on acquiring data from more individuals using the ear-worn IMU sensor and developing more reliable algorithms with better generalizability.

\section{Acknowledgement}
This work was in part supported by National Natural and Science Foundation of China under Grant 62203296, Shanghai Pujiang Program under Grant 22PJ1405500, and Science and Technology Commission of Shanghai Municipality under Grant 20DZ2220400.
\bibliographystyle{ACM-Reference-Format}
\balance
\bibliography{earcomp24}


\begin{thebibliography}{15}


\ifx \showCODEN    \undefined \def \showCODEN     #1{\unskip}     \fi
\ifx \showDOI      \undefined \def \showDOI       #1{#1}\fi
\ifx \showISBNx    \undefined \def \showISBNx     #1{\unskip}     \fi
\ifx \showISBNxiii \undefined \def \showISBNxiii  #1{\unskip}     \fi
\ifx \showISSN     \undefined \def \showISSN      #1{\unskip}     \fi
\ifx \showLCCN     \undefined \def \showLCCN      #1{\unskip}     \fi
\ifx \shownote     \undefined \def \shownote      #1{#1}          \fi
\ifx \showarticletitle \undefined \def \showarticletitle #1{#1}   \fi
\ifx \showURL      \undefined \def \showURL       {\relax}        \fi
\providecommand\bibfield[2]{#2}
\providecommand\bibinfo[2]{#2}
\providecommand\natexlab[1]{#1}
\providecommand\showeprint[2][]{arXiv:#2}

\bibitem[Atallah et~al\mbox{.}(2013)]%
        {atallah2013ear}
\bibfield{author}{\bibinfo{person}{Louis Atallah}, \bibinfo{person}{Omer Aziz}, {et~al\mbox{.}}} \bibinfo{year}{2013}\natexlab{}.
\newblock \showarticletitle{An ear-worn sensor for the detection of gait impairment after abdominal surgery}.
\newblock \bibinfo{journal}{\emph{Surgical innovation}} \bibinfo{volume}{20}, \bibinfo{number}{1} (\bibinfo{year}{2013}), \bibinfo{pages}{86--94}.
\newblock


\bibitem[Atallah et~al\mbox{.}(2011)]%
        {atallah2011sensor}
\bibfield{author}{\bibinfo{person}{Louis Atallah}, \bibinfo{person}{Benny Lo}, {et~al\mbox{.}}} \bibinfo{year}{2011}\natexlab{}.
\newblock \showarticletitle{Sensor positioning for activity recognition using wearable accelerometers}.
\newblock \bibinfo{journal}{\emph{IEEE transactions on biomedical circuits and systems}} \bibinfo{volume}{5}, \bibinfo{number}{4} (\bibinfo{year}{2011}), \bibinfo{pages}{320--329}.
\newblock


\bibitem[Aziz et~al\mbox{.}(2023)]%
        {aziz2023earables}
\bibfield{author}{\bibinfo{person}{Abdul Aziz}, \bibinfo{person}{Ravi Karkar}, {et~al\mbox{.}}} \bibinfo{year}{2023}\natexlab{}.
\newblock \showarticletitle{Earables as Medical Devices: Opportunities and Challenges}. In \bibinfo{booktitle}{\emph{Adjunct Proceedings of the 2023 ACM International Joint Conference on Pervasive and Ubiquitous Computing \& the 2023 ACM International Symposium on Wearable Computing}}. \bibinfo{pages}{339--341}.
\newblock


\bibitem[Cola et~al\mbox{.}(2015)]%
        {cola2015node}
\bibfield{author}{\bibinfo{person}{Guglielmo Cola}, \bibinfo{person}{Marco Avvenuti}, {et~al\mbox{.}}} \bibinfo{year}{2015}\natexlab{}.
\newblock \showarticletitle{An on-node processing approach for anomaly detection in gait}.
\newblock \bibinfo{journal}{\emph{IEEE Sensors Journal}} \bibinfo{volume}{15}, \bibinfo{number}{11} (\bibinfo{year}{2015}), \bibinfo{pages}{6640--6649}.
\newblock


\bibitem[Diao et~al\mbox{.}(2020)]%
        {diao2020novel}
\bibfield{author}{\bibinfo{person}{Yanan Diao}, \bibinfo{person}{Yu Ma}, {et~al\mbox{.}}} \bibinfo{year}{2020}\natexlab{}.
\newblock \showarticletitle{A novel gait parameter estimation method for healthy adults and postoperative patients with an ear-worn sensor}.
\newblock \bibinfo{journal}{\emph{Physiological measurement}} \bibinfo{volume}{41}, \bibinfo{number}{5} (\bibinfo{year}{2020}), \bibinfo{pages}{05NT01}.
\newblock


\bibitem[Guo et~al\mbox{.}(2021)]%
        {guo2021mcdcd}
\bibfield{author}{\bibinfo{person}{Yao Guo}, \bibinfo{person}{Xiao Gu}, {and} \bibinfo{person}{Guang-Zhong Yang}.} \bibinfo{year}{2021}\natexlab{}.
\newblock \showarticletitle{MCDCD: Multi-source unsupervised domain adaptation for abnormal human gait detection}.
\newblock \bibinfo{journal}{\emph{IEEE Journal of Biomedical and Health Informatics}} \bibinfo{volume}{25}, \bibinfo{number}{10} (\bibinfo{year}{2021}), \bibinfo{pages}{4017--4028}.
\newblock


\bibitem[Guo et~al\mbox{.}(2022)]%
        {guo2022detection}
\bibfield{author}{\bibinfo{person}{Yao Guo}, \bibinfo{person}{Jianxin Yang}, {et~al\mbox{.}}} \bibinfo{year}{2022}\natexlab{}.
\newblock \showarticletitle{Detection and assessment of Parkinson's disease based on gait analysis: A survey}.
\newblock \bibinfo{journal}{\emph{Frontiers in aging neuroscience}}  \bibinfo{volume}{14} (\bibinfo{year}{2022}), \bibinfo{pages}{916971}.
\newblock


\bibitem[Jarchi et~al\mbox{.}(2014)]%
        {jarchi2014validation}
\bibfield{author}{\bibinfo{person}{Delaram Jarchi}, \bibinfo{person}{Benny Lo}, {et~al\mbox{.}}} \bibinfo{year}{2014}\natexlab{}.
\newblock \showarticletitle{Validation of the e-AR sensor for gait event detection using the parotec foot insole with application to post-operative recovery monitoring}. In \bibinfo{booktitle}{\emph{2014 11th International Conference on Wearable and Implantable Body Sensor Networks}}. IEEE, \bibinfo{pages}{127--131}.
\newblock


\bibitem[Jiang et~al\mbox{.}(2017)]%
        {jiang2017robust}
\bibfield{author}{\bibinfo{person}{Shuo Jiang}, \bibinfo{person}{Xingchen Wang}, \bibinfo{person}{Maria Kyrarini}, {and} \bibinfo{person}{Axel Gr{\"a}ser}.} \bibinfo{year}{2017}\natexlab{}.
\newblock \showarticletitle{A robust algorithm for gait cycle segmentation}. In \bibinfo{booktitle}{\emph{2017 25th european signal processing conference (eusipco)}}. IEEE, \bibinfo{pages}{31--35}.
\newblock


\bibitem[Kawsar et~al\mbox{.}(2018)]%
        {kawsar2018earables}
\bibfield{author}{\bibinfo{person}{Fahim Kawsar}, \bibinfo{person}{Chulhong Min}, {et~al\mbox{.}}} \bibinfo{year}{2018}\natexlab{}.
\newblock \showarticletitle{Earables for personal-scale behavior analytics}.
\newblock \bibinfo{journal}{\emph{IEEE Pervasive Computing}} \bibinfo{volume}{17}, \bibinfo{number}{3} (\bibinfo{year}{2018}), \bibinfo{pages}{83--89}.
\newblock


\bibitem[Mazilu et~al\mbox{.}(2013)]%
        {mazilu2013engineers}
\bibfield{author}{\bibinfo{person}{Sinziana Mazilu}, \bibinfo{person}{Ulf Blanke}, {et~al\mbox{.}}} \bibinfo{year}{2013}\natexlab{}.
\newblock \showarticletitle{Engineers meet clinicians: augmenting Parkinson's disease patients to gather information for gait rehabilitation}. In \bibinfo{booktitle}{\emph{Proceedings of the 4th augmented human international conference}}. \bibinfo{pages}{124--127}.
\newblock


\bibitem[Min et~al\mbox{.}(2018)]%
        {min2018exploring}
\bibfield{author}{\bibinfo{person}{Chulhong Min}, \bibinfo{person}{Akhil Mathur}, {and} \bibinfo{person}{Fahim Kawsar}.} \bibinfo{year}{2018}\natexlab{}.
\newblock \showarticletitle{Exploring audio and kinetic sensing on earable devices}. In \bibinfo{booktitle}{\emph{Proceedings of the 4th ACM Workshop on Wearable Systems and Applications}}. \bibinfo{pages}{5--10}.
\newblock


\bibitem[Prakash et~al\mbox{.}(2019)]%
        {prakash2019stear}
\bibfield{author}{\bibinfo{person}{Jay Prakash}, \bibinfo{person}{Zhijian Yang}, {et~al\mbox{.}}} \bibinfo{year}{2019}\natexlab{}.
\newblock \showarticletitle{Stear: Robust step counting from earables}. In \bibinfo{booktitle}{\emph{Proceedings of the 1st International Workshop on Earable Computing}}. \bibinfo{pages}{36--41}.
\newblock


\bibitem[Schuster and Paliwal(1997)]%
        {schuster1997bidirectional}
\bibfield{author}{\bibinfo{person}{Mike Schuster} {and} \bibinfo{person}{Kuldip~K Paliwal}.} \bibinfo{year}{1997}\natexlab{}.
\newblock \showarticletitle{Bidirectional recurrent neural networks}.
\newblock \bibinfo{journal}{\emph{IEEE transactions on Signal Processing}} \bibinfo{volume}{45}, \bibinfo{number}{11} (\bibinfo{year}{1997}), \bibinfo{pages}{2673--2681}.
\newblock


\bibitem[Zhou et~al\mbox{.}(2020)]%
        {zhou2020validation}
\bibfield{author}{\bibinfo{person}{Lin Zhou}, \bibinfo{person}{Can Tunca}, {et~al\mbox{.}}} \bibinfo{year}{2020}\natexlab{}.
\newblock \showarticletitle{Validation of an IMU gait analysis algorithm for gait monitoring in daily life situations}. In \bibinfo{booktitle}{\emph{2020 42nd Annual International Conference of the IEEE Engineering in Medicine \& Biology Society (EMBC)}}. IEEE, \bibinfo{pages}{4229--4232}.
\newblock


\end{thebibliography}


\end{document}